\documentclass[final,5p,double-column]{elsarticle}

\usepackage{lineno,hyperref}

\usepackage{graphicx}
\usepackage{natbib}
\usepackage{amsmath}
\usepackage{amsfonts}
\usepackage{xcolor}
\usepackage{graphicx}
\usepackage{overpic}
\usepackage{subfig}

\journal{Journal of \LaTeX\ Templates}

\begin{document}

\begin{frontmatter}

\title{3D Hand Pose and Shape Estimation from RGB Images for Keypoint-Based Hand Gesture Recognition}

\author[sapienza]{Danilo Avola}
\author[sapienza]{Luigi Cinque}
\author[sapienza]{Alessio Fagioli}
\author[udine]{Gian Luca Foresti}
\author[bristol]{\\Adriano Fragomeni}
\author[sapienza]{Daniele Pannone}
\cortext[corrauth]{Corresponding author email: fagioli@di.uniroma1.it}

\address[sapienza]{Department
of Computer Science, Sapienza University,\\Via Salaria 113, Rome, 00198, Italy}
\address[udine]{Department of Mathematics, Computer Science and Physics, University of Udine,\\Via delle Scienze 20, Udine, 33100, Italy}
\address[bristol]{Department of Computer Science, University of Bristol,\\Merchant Venturers Building Woodland Road, BS8 1UB, Bristol, UK}

\begin{abstract}
Estimating the 3D pose of a hand from a 2D image is a well-studied problem and a requirement for several real-life applications such as virtual reality, augmented reality, and hand gesture recognition. Currently, reasonable estimations can be computed from single RGB images, especially when a multi-task learning approach is used to force the system to consider the shape of the hand when its pose is determined. However, depending on the method used to represent the hand, the performance can drop considerably in real-life tasks, suggesting that stable descriptions are required to achieve satisfactory results. In this paper, we present a keypoint-based end-to-end framework for 3D hand and pose estimation and successfully apply it to the task of hand gesture recognition as a study case. Specifically, after a pre-processing step in which the images are normalized, the proposed pipeline uses a multi-task semantic feature extractor generating 2D heatmaps and hand silhouettes from RGB images, a viewpoint encoder to predict the hand and camera view parameters, a stable hand estimator to produce the 3D hand pose and shape, and a loss function to guide all of the components jointly during the learning phase. Tests were performed on a 3D pose and shape estimation benchmark dataset to assess the proposed framework, which obtained state-of-the-art performance. Our system was also evaluated on two hand-gesture recognition benchmark datasets and significantly outperformed other keypoint-based approaches, indicating that it is an effective solution that is able to generate stable 3D estimates for hand pose and shape.
\end{abstract}


\end{frontmatter}


\section{Introduction}\label{intro}
3D pose estimation is a fundamental technology that has become very important in recent years for computer vision-based tasks, primarily due to advances in several practical applications such as virtual reality (VR) \citep{alam2022unified}, augmented reality (AR) \citep{makar2014interframe}, sign language recognition \citep{avola2018exploiting} and, more generally, gesture recognition \citep{guo2021normalized}. In these fields, much of the current effort is directed towards the pose estimation of hands since, due to the high number of joints, they are one of the most complex components of the human body \citep{rehg1994visual}. To address this complexity, researchers generally follow either a model-driven or data-driven strategy. The former uses articulated hand models to describe bones, muscles, and tendons, for example, through kinematics constraints \citep{de2010variational,de2011model}, whereas the latter directly exploits depth, RGB-D, or RGB images \citep{zhao2017simple,dibra2018monocular,mueller2018ganerated} to extract keypoints that represent a hand. Although both strategies have their merits, data-driven approaches have allowed various systems to achieve significant performance while remaining more straightforward to implement and are usually the preferred choice between the two strategies. 
Although early data-driven works were based on machine learning or computer vision algorithms such as random forest \citep{keskin2012hand} and geodesic distance-based systems \citep{tang2014latent}, attention has recently shifted towards deep learning methods \citep{li2019survey}. This is due to the high performance obtained in a heterogeneous range of fields such as emotion recognition \citep{sheng2021multi,avola2020deep}, medical image analysis \citep{yan2021development,avola2021multimodal}, and person re-identification \citep{prasad2021spatio,wu2022learning}, as well as the availability of commodity hardware for capture systems \citep{yuan2017bighand2} that can provide different types of input (e.g., depth maps). 

For the task of 3D hand pose estimation, methods based on deep learning architecture configurations such as multilayer perceptrons (MLPs), convolutional neural networks (CNNs), and autoencoders have been proposed. These methods usually analyze hand keypoints via 2D heatmaps, which represent hand skeletons and are extrapolated from depth \citep{zhao2017simple}, RGB-D \citep{dibra2018monocular}, or RGB \citep{iqbal2018hand} input images. While the first two of these input options provide useful information for estimating the 3D pose through the depth component, near state-of-the-art results have been obtained by exploiting single RGB images \citep{cai2018weakly}. 
The reasons for this are twofold. Firstly, even though commodity sensors are available, it is hard to acquire and correctly label a depth dataset due to the intrinsic complexity of the hand skeleton, which has resulted in a lack of such datasets. Secondly, RGB images can easily be exploited in conjunction with data augmentation strategies, thus allowing a network to be trained more easily \citep{tanner1987calculation}. 
To further improve the estimation of 3D hand pose from 2D images, several recent works have exploited more complex architectures (e.g., residual network) and the multi-task learning paradigm by generating 3D hand shapes together with their estimated pose. 
By leveraging these approaches together with hand models, such as the model with articulated and non-rigid deformations (MANO) \citep{romero2017embodied} and graph CNNs \citep{ge20193d}, various systems have achieved state-of-the-art performance. In particular, they can produce correct hand shapes and obtain more accurate pose estimations from an analysis of 2D hand heatmaps and depth maps generated from a single RGB input image \citep{zhang2019end}.

Inspired by the results reported in other works, we propose a keypoint-based end-to-end framework that can give state-of-the-art performance for both 3D hand pose and shape estimation; we also show that this system can be successfully applied to the task of hand gesture recognition and can outperform other keypoint-based works.
In more detail, our framework first applies a pre-processing phase to normalize RGB images containing hands. 
A semantic feature extractor (SFE) with a multi-task stacked hourglass network is employed for the first time in the literature to simultaneously generate 2D heatmaps and hand silhouettes starting from an RGB image. 
A novel viewpoint encoder (VE) is used to reduce the number of parameters required to encode the feature space representing the camera view during the computation of the viewpoint vector. 
A stable hand pose/shape estimator (HE) based on a fine-tuned MANO layer is employed in conjunction with an improved version of a neural 3D mesh renderer \citep{kato2018neural}. This is extended via a custom weak perspective projection through the 2D re-projection of the generated 3D joint positions and meshes. Finally, a multi-task loss function is used to train the various framework components to carry out the 3D hand pose and shape estimation. 

The main contributions of this paper can be summarized as follows:
\begin{itemize}
    \item We present a comprehensive end-to-end framework based on keypoints that combines and improves upon several different technologies to generate 3D hand pose and shape estimations;
    \item We propose a multi-task SFE, design an optimized VE, and introduce a re-projection procedure for more stable outputs;
    \item We evaluate the generalization of the capabilities of our model to the task of hand gesture recognition and show that it outperforms other relevant keypoint-based approaches developed for 3D hand estimation.
\end{itemize}

The rest of this paper is organized as follows. Section~\ref{sec:related} introduces relevant work that inspired this study. Section~\ref{sec:method} presents an exhaustive description of the components of our framework. Section~\ref{sec:results} describes the experiments performed to validate the proposed approach and presents a comparison with other state-of-the-art methods for hand pose, shape estimation, and hand-gesture recognition tasks. Finally, Section~\ref{sec:conclusions} draws some conclusions from this study.

\begin{figure}[t!]
    \centering
    \includegraphics[width=\columnwidth]{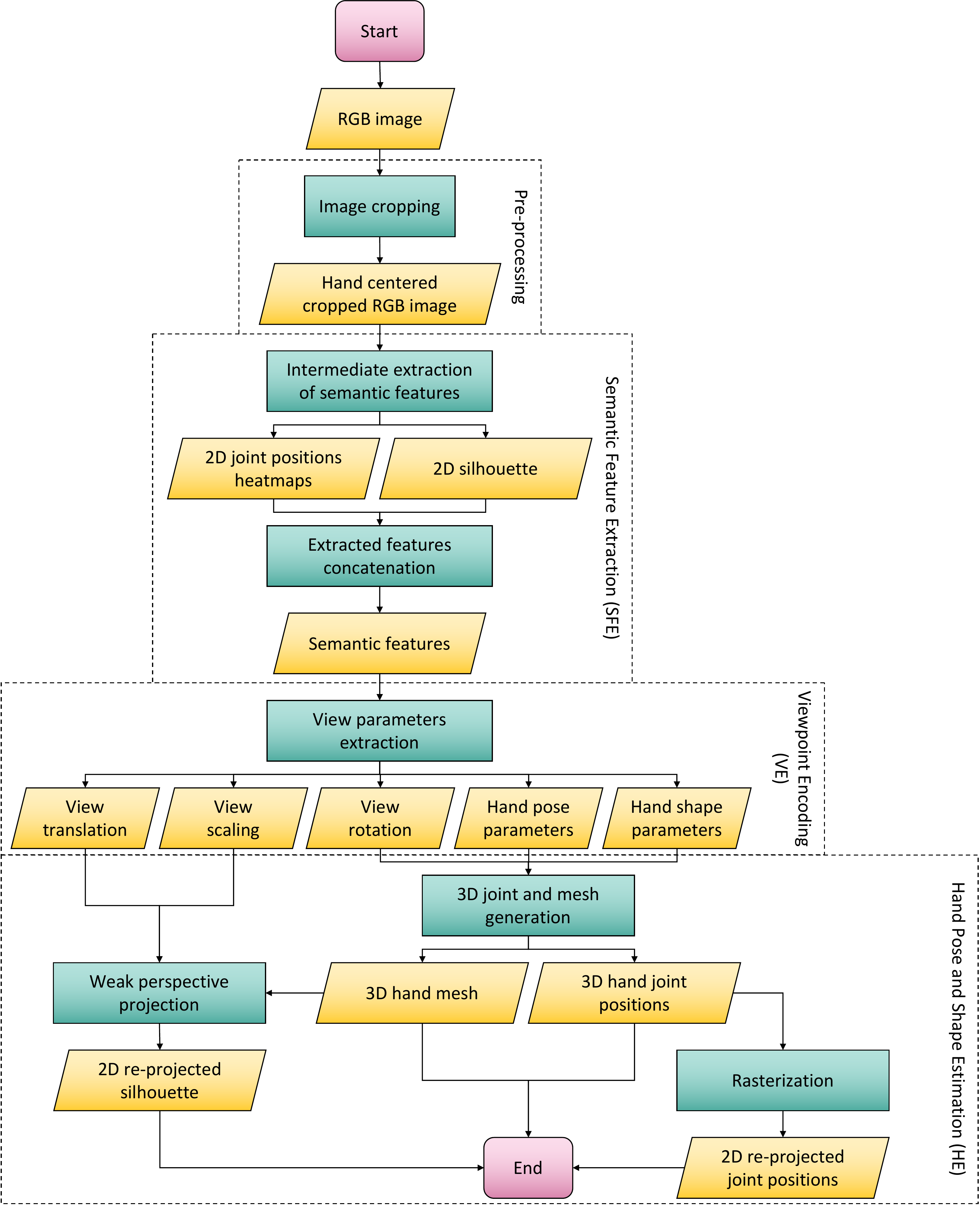}
    \caption{Proposed framework flowchart.}
    \label{fig:flowchart}
\end{figure}

\section{Related Work}\label{sec:related}
Methods of 3D pose and shape estimation generally exploit depth, RGB-D, or RGB images. The last of these is usually the preferred solution due to the availability of datasets; however, approaches that leverage depth information have provided solutions and ideas that can also be applied to standalone RGB images.
For instance, the study in \cite{sun2015cascaded} introduces a hierarchical regression algorithm that starts with depth maps and describes hand keypoints based on their geometric properties and divides the hand into meaningful components such as the fingers and palm to obtain 3D poses. 
In contrast, the study in \cite{malik2020handvoxnet} uses depth maps to build both 3D hand shapes and surfaces by defining 3D voxelized depth maps that can mitigate possible depth artifacts. 
The representation of meaningful hand components and reductions in input noise are also relevant problems for RGB and RGB-D images. 
For example, when considering RGB-D inputs for the task of 3D hand pose estimation, \cite{oikonomidis2011efficient} and \cite{qian2014realtime} use the depth component to define hand characteristics through geometric primitives that are later matched with the RGB information to generate 3D poses, and hence to track the hands. 
Specifically, spheres, cones, cylinders, and ellipsoids are used to describe the palm and fingers in \cite{oikonomidis2011efficient}, while the approach in \cite{qian2014realtime} employs only sphere primitives for faster computation.
Using a different technique, \cite{dibra2018monocular} focuses on handling input noise by using synthetic RGB-D images to train a CNN. In particular, the use of artificial RGB/depth image pairs is shown by the authors to alleviate the effects of missing or unlabeled depth datasets.

\begin{figure*}[t]
    \centering
    \includegraphics[width=\textwidth]{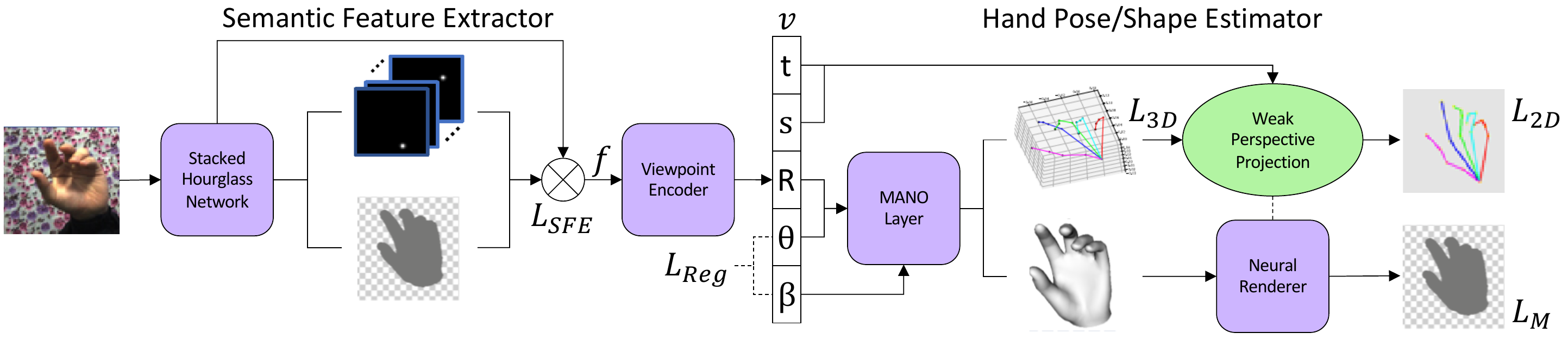}
    \caption{Proposed framework architecture overview.}
    \label{fig:archi}
\end{figure*}

Approaches that exploit depth information are inherently more suitable for the 3D pose estimation task since they suffer less from image ambiguities compared to systems based exclusively on RGB images. 
In view of this, \cite{iqbal2018hand} introduces a 2.5D representation by building a latent depth space via an autoencoder that retains depth information without directly using this type of data. The authors further refine this latent space through an element-wise multiplication with 2D heatmaps to increase the depth consistency and obtain realistic 3D hand poses. 
Another work that focuses on depth retention without using the extra information at test time is presented in \cite{cai20203d}. This scheme first employs a conditional variational autoencoder (VAE) to build the latent distribution of joints via the 2D heatmaps extracted from the input RGB image, and then exploits a weak-supervision approach through a depth regularizer that forces the autoencoder to consider automatically generated depth information in its latent space at training time. 
A similar weak-supervision rationale is also applied in \cite{zhang2019end} where, in addition to depth information, the hand shape consistency is evaluated through a neural renderer. 
More specifically, by exploiting the outputs of the MANO layer (i.e., the 3D hand pose and mesh), the authors project the 3D joint coordinates defining the hand pose into a 2D space to account for depth information, and implement a neural renderer to generate silhouettes from hand shapes to increase the consistency of the results. In the present work, we refine this procedure further via a weak re-projection that is applied to both 3D joint locations and mesh so that the proposed framework can also be applied to a different task.

Accounting for depth information without directly using such data enables higher performance when only RGB images are analyzed. 
However, this image format introduces several challenges that must be addressed, such as different camera view parameters, background clutter, occlusions, and hand segmentation.
In general, to handle these problems, RGB-based methods define a pipeline that includes feature extraction from the input image (usually in the form of 2D heatmaps), a latent space representation of such features to allow for the extrapolation of meaningful view parameters, and the 3D hand pose estimation based on the computed view parameters. 
For instance, the study in \cite{zimmermann2017learning} implements a CNN called HandSegNet to identify hand silhouettes so that the input images can be cropped and resized around the hand. A second CNN (PoseNet) is then used to extract the features, i.e., 2D heatmaps, allowing the network to estimate the 3D pose via symmetric streams and to analyze the prior pose and the latent pose representation derived by the network.
The authors of \cite{baek2020weakly} instead devise a domain adaptation strategy in which a generative adversarial network (GAN), driven by the 2D heatmaps extracted from the input by a convolutional pose machine, automatically outputs hand-only images from hand-object images. The resulting hand-only images are then used to estimate the correct 3D pose, even in the case of occlusions (e.g., from the object being held). Object shapes are also exploited in \cite{hasson2019learning} to handle occlusions that arise during the task of 3D hand pose estimation. The authors describe the use of two parallel encoders to obtain latent representations of both hand and object, which are in turn employed to define meaningful hand-object constellations via a custom contact loss, so that consistent 3D hand poses can be generated. In contrast, the authors of \cite{yang2019disentangling} directly address the issues of background clutter and different camera view parameters by designing a disentangling VAE (dVAE) to decouple hand, background, and camera view, using a latent variable model, so that a MANO layer can receive the correct input to generate the 3D hand pose. 
Furthermore, through the use of the dVAE, the authors are able to synthesize realistic hand images in a given 3D pose, which may also alleviate the issue of low numbers of available datasets. 

The use of the pipeline described above allows models to achieve state-of-the-art performance on the 3D hand pose estimation task. Nevertheless, depending on the 3D joint position generation procedure, a good latent space representation is necessary to ensure an effective system. For example, the scheme in \cite{ge20193d} utilizes a stacked hourglass to retrieve 2D heatmaps from the input RGB image. A residual network is then implemented to generate a meaningful latent space, and a graph CNN is built to define both the 3D pose and the shape of the input hand. To further improve the results, the authors pre-train all networks on a synthetic dataset before fine-tuning them on the estimation task. 
Unlike the model introduced in \cite{ge20193d}, which is used as a starting point, the framework presented here extends the use of a stacked hourglass and residual network by defining a multi-task SFE and VE, respectively. Moreover, only the multi-task feature extractor is pre-trained on synthetic data so that the VE is able to obtain a hand abstraction that can be utilized in different tasks such as hand gesture recognition. 

Finally, the latent space representation is also relevant when using other procedures for 3D pose estimation, such as the MANO layer, as discussed in \cite{boukhayma20193d} and \cite{baek2019pushing}. In more detail, the former of these schemes extracts 2D heatmaps via a CNN, then employs an encoder to generate the MANO layer view parameters directly, while in the latter approach, a latent space is built by an evidence estimator module that leverages a convolutional pose machine, which generates the required parameters for the MANO layer. Both methods implement a re-projection procedure for the 3D joints, and this was further extended in \cite{baek2019pushing} via iterative re-projection to improve the final 3D hand pose and hence the shape estimations. In the framework presented here, we exploit this interesting strategy in a different way from the two approaches described above by also applying the re-projection procedure to the mesh generated by the MANO layer so that the estimation can benefit from both outputs rather than only the 3D locations.

\section{Method}\label{sec:method}
The proposed framework for 3D hand pose and shape estimation starts from a pre-processed hand image input, and first generates 2D heatmaps and hand silhouettes through the use of a multi-task SFE. Secondly, it estimates the camera, hand pose, and hand shape view parameters by exploiting the semantic features using a VE. Finally, it computes the hand 2D and 3D joints, mesh and silhouette by feeding the estimated view parameters to the HE, which consists of a MANO layer, weak perspective projection, and neural renderer components. We note that a single compound loss function is employed to drive the learning phase of the various modules jointly. 
A flowchart for the framework is shown in Fig.~\ref{fig:flowchart} and the high-level pipeline is illustrated in Fig.~\ref{fig:archi}.
 
\begin{figure*}[t]
    \centering
    \includegraphics[width=\textwidth]{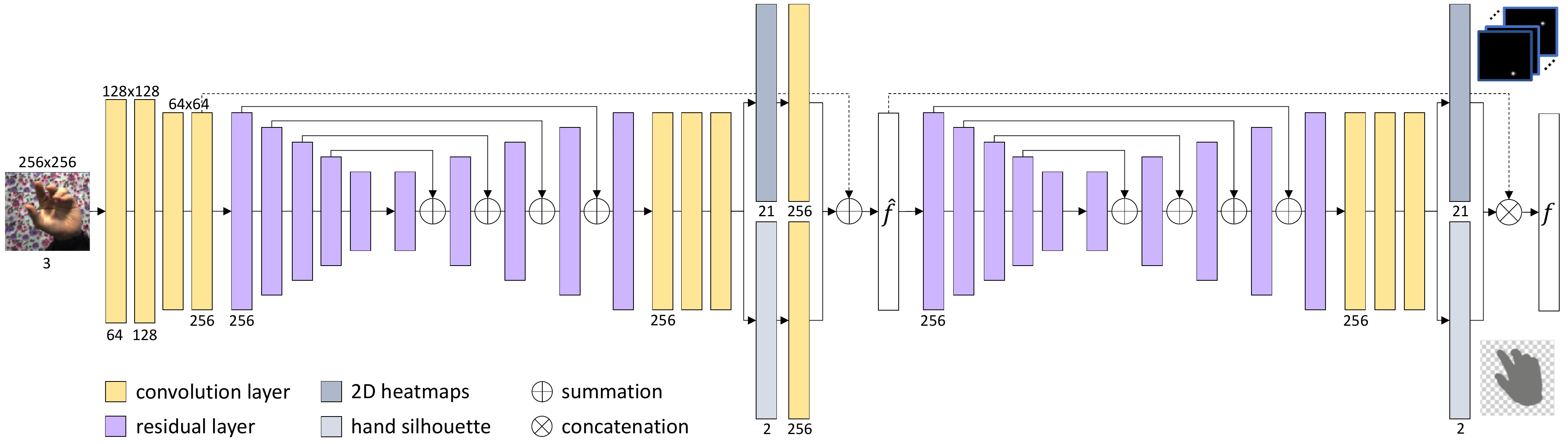}
    \caption{Modified stacked hourglass network used as a multi-task semantic feature extractor.}
    \label{fig:sfe}
\end{figure*}

\subsection{Pre-processing}
A necessary step in order to handle different image sizes and to reduce the amount of background clutter in the samples in a given dataset is pre-processing of the input image. More specifically, to allow the proposed framework to focus on hands, each image is modified so that the hand is always centered and there is as little background as possible, while still retaining all of the 21 hand joints keypoints. The hand is centered by selecting the metacarpophalangeal joint (i.e., the base knuckle) of the middle finger as a center crop point $p_c$. The crop size $l$ is then computed as follows:
\begin{equation}
    l = 2*\max((p_{max}-p_c), (p_c-p_{min})),
\end{equation}
where $p_{max}$ and $p_{min}$ are the joint keypoint coordinates with the largest and smallest $(x, y)$ distance with respect to $p_c$. $l$ is enlarged by another $20\%$ (i.e., $\sim$20 px padding in all directions) to ensure all hand joints are fully visible inside the cropped area.
 
\subsection{Semantic Feature Extractor}
Inspired by the results obtained in \cite{ge20193d}, we implemented a modified version of a stacked hourglass network \citep{newell2016stacked} to take advantage of the multi-task learning approach. In particular, 2D heatmaps and hand silhouette estimates are generated based on a $256\times256\times3$ normalized (i.e., with zero-mean and unit variance) image $I_{normalized}$. The hourglass architecture was selected as it can capture many features, such as hand orientation, articulation structure, and joint relationships, by analyzing the input image at different scales. Four convolutional layers are employed in the proposed architecture to reduce the input image to a size of $64\times64$ via two max pooling operations in the first and third layers. The downsized images are then fed to the hourglass module and intermediate heatmaps and silhouettes are generated by processing local and global contexts in a multi-task learning scenario.
These two outputs, of size $64\times64\times21$ (i.e., one channel per hand joint) and $64\times64\times2$ (i.e., back and foreground channels) for 2D heatmaps and silhouette, respectively, are then mapped to a larger number of channels via a $1\times1$ convolution to reintegrate the intermediate feature predictions into the feature space. 
These representations are then summed with the hourglass input into a single vector $\hat{f}$, thus effectively introducing long skip connections to reduce data loss for the second hourglass module. Finally, this second module is employed to extract the semantic feature vector $f$ that contains the effective 2D heatmaps and hand silhouette used by the VE to regress camera view, hand pose, and hand shape parameters. Note that unlike $\hat{f}$, the vector $f$ is computed via concatenation of 2D heatmaps, hand silhouette, and $\hat{f}$, which provides the VE with a comprehensive representation of the input. 

In the hourglass components, each layer employs two residual modules \citep{he2016deep} for both downsampling and upsampling sequences. In the former, the features are processed down to a very low resolution (i.e., $4\times4$), whereas in the latter, images are upsampled and combined with the extracted features across all scales. 
In more detail, the information on two adjacent resolutions is merged by first increasing the lower resolution through the nearest neighbor interpolation and then performing an element-wise addition. After the upsampling sequence, three consecutive $1\times1$ convolutions are applied to obtain the multi-task output (i.e., 2D heatmaps and silhouette) used by the VE. The presented SFE architecture with the corresponding layer sizes are illustrated in Fig.~\ref{fig:sfe}.

\begin{figure}[t]
    \centering
    \includegraphics[width=\columnwidth]{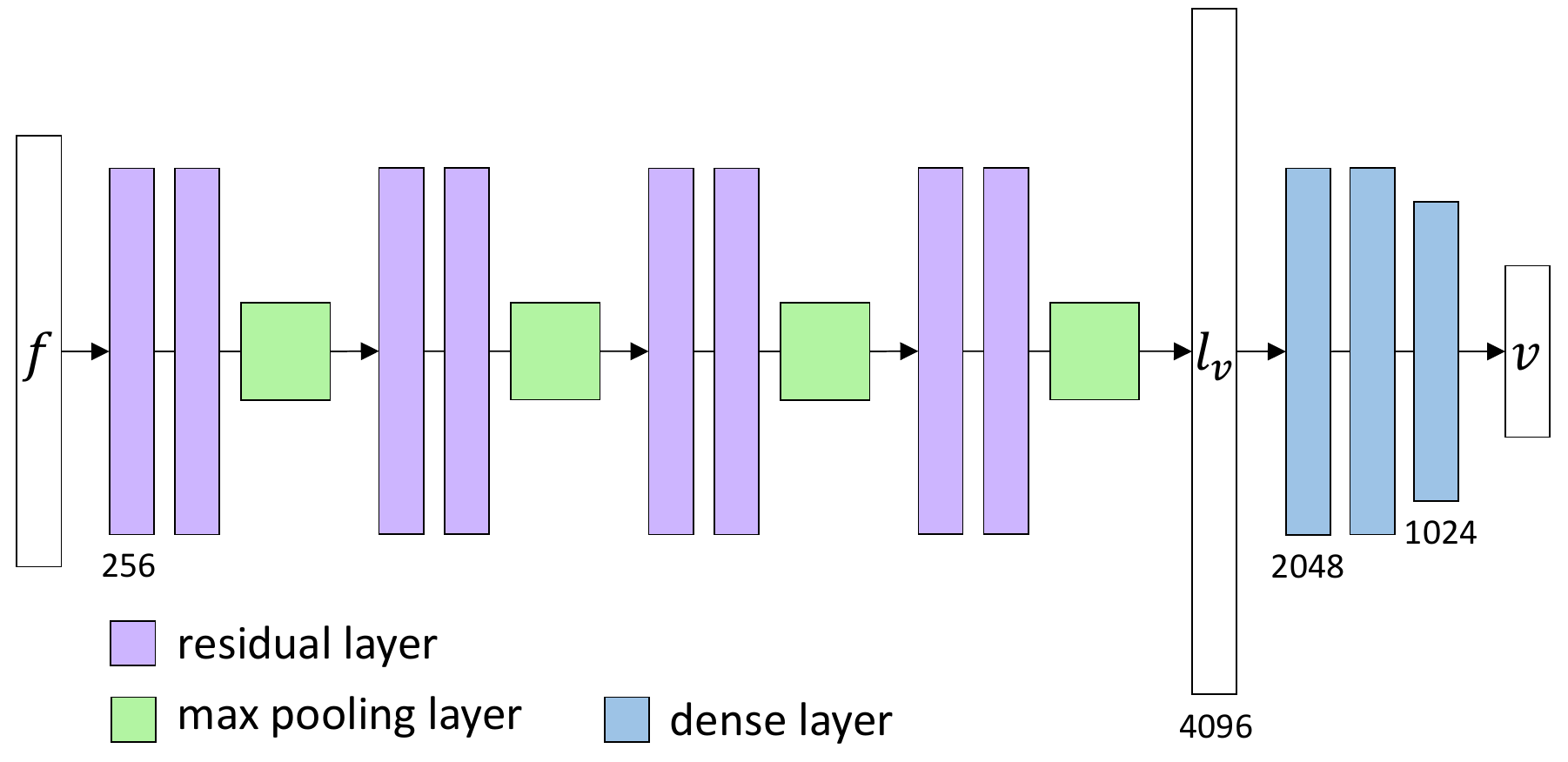}
    \caption{Viewpoint encoder architecture.}
    \label{fig:wpe}
\end{figure}

\subsection{Viewpoint Encoder}
The extracted semantic feature vector $f$ is resized to dimensions of $256\times256$ and is then used as input for the second framework component, i.e., the VE, which has two main objectives. Firstly, this unit generates a set of parameters $v$, which are employed by the last component of the pipeline to produce the 3D hand pose and shape. 
The vector $v$ contains the camera view translation $t\in \mathbb{R}^2$, scaling $s\in \mathbb{R}^+$, and rotation $R \in \mathbb{R}^3$, as well as the hand pose $\theta\in \mathbb{R}^{45}$ and shape $\beta \in \mathbb{R}^{10}$ values necessary to move from a 2D space to a 3D one. 
Secondly, the VE also needs to sensibly reduce the number of trainable parameters of the architecture to satisfy the hardware constraints. 
In more detail, given the semantic feature vector $f$, a flattened latent viewpoint feature space $l_v$ that encodes semantic information is obtained by using four abstraction blocks, each containing two residual layers \citep{he2016deep}, to analyze the input, and a max-pooling layer, to both consolidate the viewpoint representation and reduce the input vector down to a $64\times64$ size. 
Subsequently, to obtain the set $v=\{t, s, R, \theta, \beta\}$, the VE transforms the latent space representation $l_v$ into $v\in\mathbb{R}^{61}$, by employing three dense layers to elaborate on the representation derived by the residual layers.
We note that by employing a max-pooling layer inside each abstraction block, a smaller latent space is obtained, meaning that a lower amount of parameters needs to be trained; hence, both of the objectives for this module are achieved. 
The architecture of the VE is shown in Fig.~\ref{fig:wpe}.

\subsection{Hand Pose/Shape Estimator}
The last component of the framework utilizes the parameter set $v=\left\{t,s,R,\theta,\beta\right\}$ to generate the hand pose and shape. The estimator applies a MANO layer for the 3D joint positions and hand mesh generation. These outputs are then improved during training by leveraging a weak perspective projection procedure and a neural renderer for more accurate estimations.

\subsubsection{MANO Layer}
This layer models the properties of the hand, such as the slenderness of the fingers, the thickness of the palm, as well as the hand pose, and controls the 3D surface deformation defined from articulations. More formally, given the pose $\theta$ and shape $\beta$ parameters, the MANO hand model $M$ is defined as follows:
\begin{equation}
    M(\beta,\theta)= W(T_p(\beta,\theta),J(\beta),\theta,\mathcal{W}),
\end{equation}
where $W$ is a linear blend skinning (LBS) function \citep{loper2015smpl}; $T_p$ corresponds to the articulated mesh template to blend, consisting of $K$ joints; $J$ represents the joint locations learned from the mesh vertices via a sparse linear regressor; and $\mathcal{W}$ indicates the blend weights.

To avoid the common problems of LBS models, such as overly smooth outputs or mesh collapse near joints, the template $T_p$ is obtained by deforming a mean mesh $\hat{T}$ using the pose and blend functions $B_S$ and $B_P$, using the following equation:
\begin{equation}
    T_p = \hat{T} + B_S(\beta) + B_P(\theta),
\end{equation}
where $B_S$ and $B_P$ allow us to vary the hand shape and to capture deformations derived from bent joints, respectively, and are computed as follows:
\begin{equation}
    B_S(\beta) = \sum_{n=1}^{|\beta|}\beta_nS_n,
\end{equation}
\begin{equation}
    B_P(\theta) = \sum_{n=1}^{9K}\left(R_n(\theta)-R_n(\theta^*)\right)P_n,
\end{equation}
where $S_n\in S$ are the blend shapes computed by applying principal component analysis (PCA) to a set of registered hand shapes, normalized by a zero pose $\theta^*$; $9K$ represents the rotation matrix scalars for each of the $K$ hand articulations; $R_n$ indicates the $n$-th element rotation matrix coefficients; while $P_n\in P$ corresponds to the blend poses. 
Finally, there is a natural variability among hand shapes in a human population, meaning that possible skeleton mismatches might be found in the MANO layer 3D joint output. To address this issue, we implemented a skeleton adaptation procedure following the solution presented in \cite{hasson2019learning}. Skeleton adaptation is achieved via a linear layer initialized to the identity function, which maps the MANO joints to the final joint annotations.

\subsubsection{Weak Perspective Projection} 
The weak perspective projection procedure takes as inputs the translation vector $t$, the scalar parameter $s$, and the 3D hand pose $RJ(\beta,\theta)$ derived from the MANO model $M(\theta,\beta)$, and re-projects the generated 3D keypoints back onto a 2D space to allow for identification of the 2D hand joint locations. This approximation allows us to train the model without defining intrinsic camera parameters, since consistency between the input and the projected 2D locations can be enforced, thus avoiding issues arising from the different devices calibrations that are typical of different datasets. Formally, the re-projection $w$ is computed as follows:
\begin{equation}
    w_{2D} = s\Pi(RJ(\beta,\theta)) + t,
\end{equation}
where $\Pi$ corresponds to the orthographic projection.

\begin{figure*}[t]
	\subfloat[$vi=(x_i,y_i)$ is a face vertex. $I_j$ is pixel $P_j$ color. $x_0$ is the current $x_i$ position. $x_1$ is the $x_i$ position when a face edge collides with $P_j$ center, with $x_i$ moving to its right. When $x_i=x_1$, then $I_j=I_{ij}$.]{\includegraphics[width=.485\textwidth]{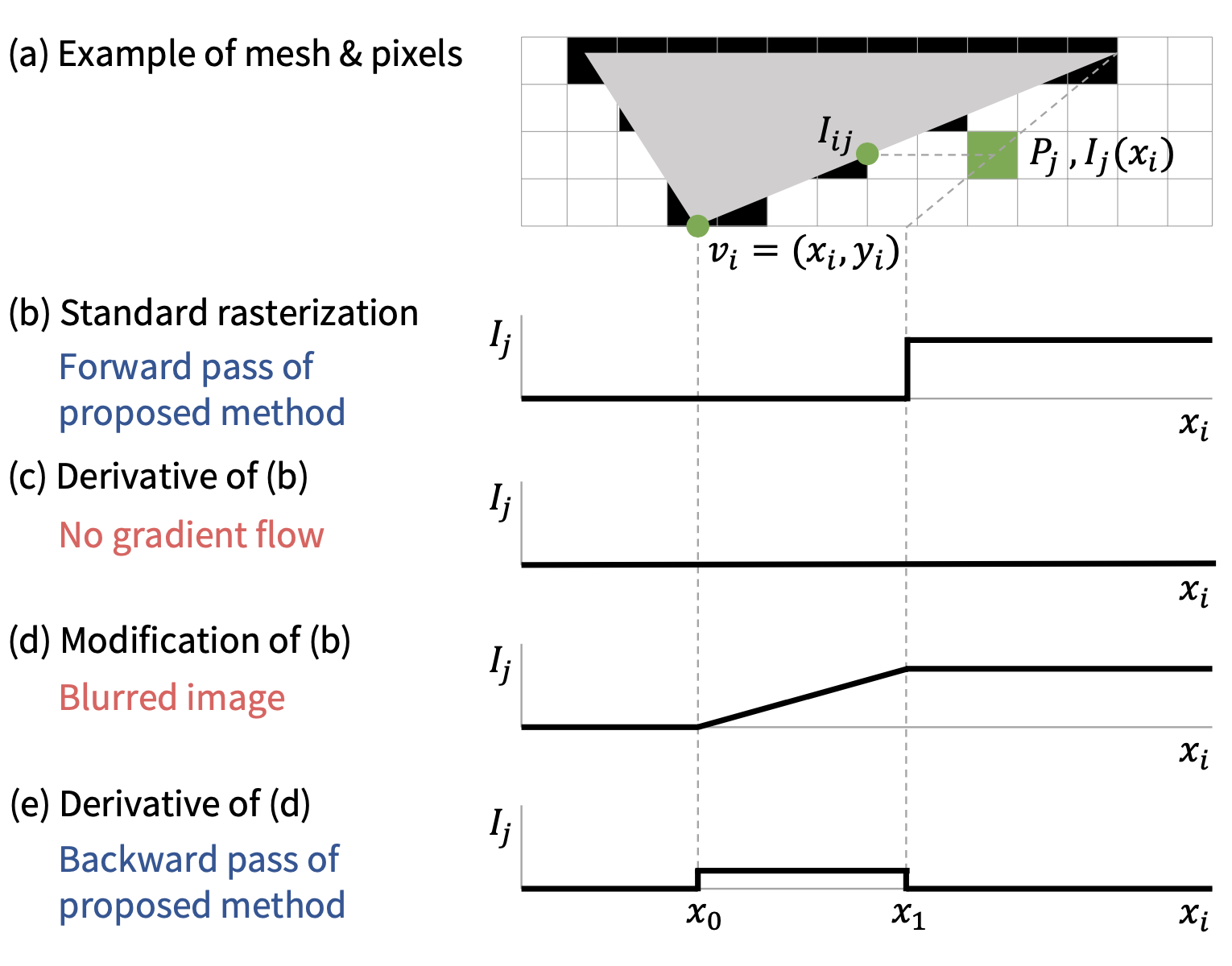}}
	\hfill
	\subfloat[$vi=(x_i,y_i)$ is a face vertex. $I_j$ is pixel $P_j$ color. $x_0$ is the current $x_i$ position. $x_1^a$ and $x_1^b$ are the $x_i$ positions when a face edge collides with $P_j$ center, with $x_i$ moving to its left or right. When $x_i=x_1^a|x_1^b$, then $I_j=I_{ij}^a|I_{ij}^b$.]{\includegraphics[width=.485\textwidth]{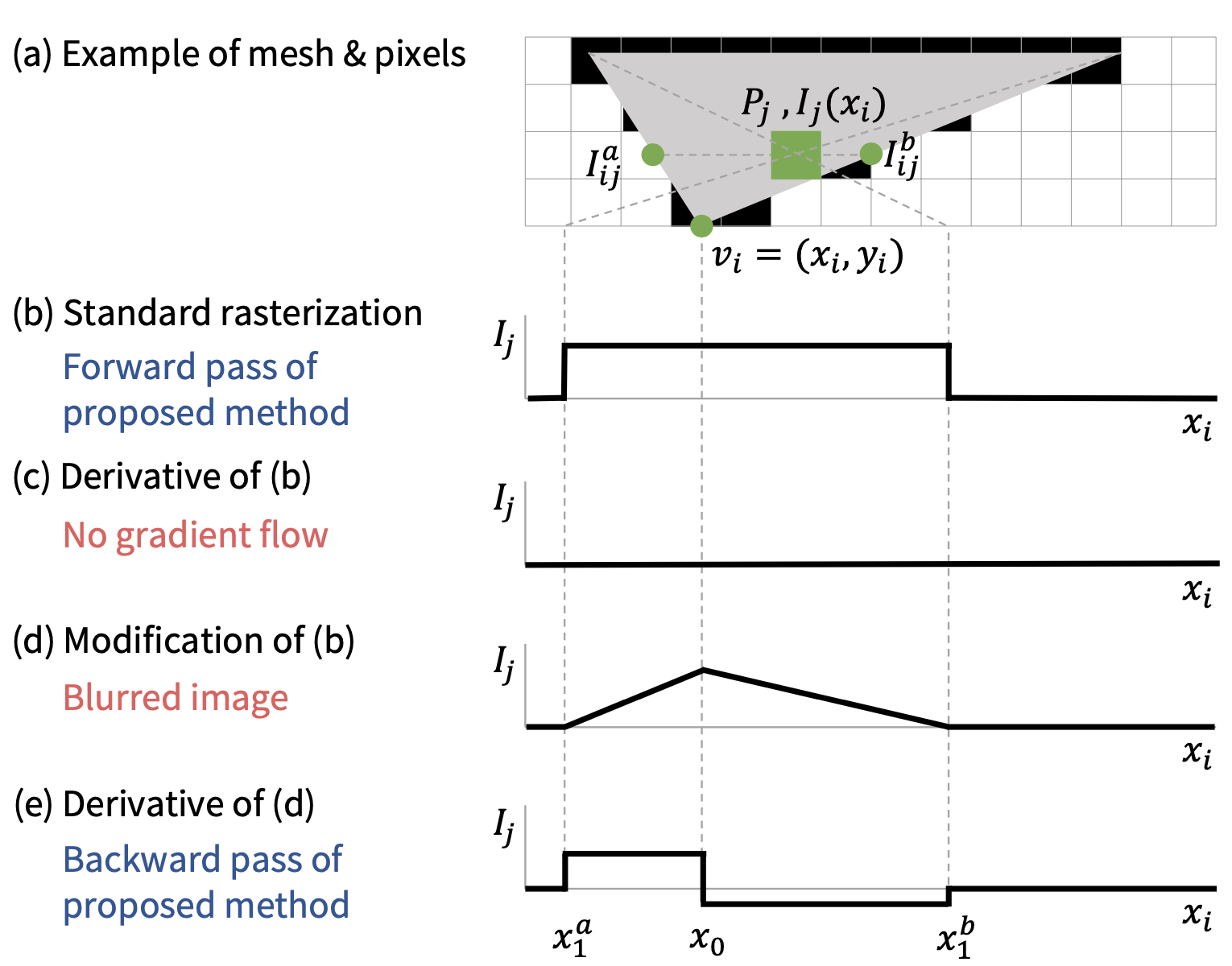}}
	\caption{Rasterization process for pixels residing outside a given face (i) and inside it (ii). Image courtesy of \cite{kato2018neural}.}
	\label{fig:rasterization}
\end{figure*}

\subsubsection{Neural Renderer}
To improve the mesh generated by the MANO layer, the differentiable neural renderer devised in \cite{kato2018neural}, which is trainable via back-propagation, is employed to rasterize the 3D shape into a hand silhouette. This silhouette, which is similar to the re-projected hand joint coordinates, is then used to improve the mesh generation. Formally, given a 3D mesh composed of $N=778$ vertices $\{v_1^o, v_2^o, \dots, v_N^o\}$, with $v_i^o\in\mathbb{R}^{3}$, and $M=1538$ faces $\{f_1, f_2, \dots, f_M \} $, with $f_j\in\mathbb{N}^{3}$, the vertices are first projected onto the 2D screen space using the weak perspective projection via the following equation:
\begin{equation}
    v_i^s = s\Pi(Rv_i^o) + t,
\end{equation}
where $R$ corresponds to the rotation matrix used to build the MANO hand model $M(\beta,\theta)$. Rasterization is then applied to generate an image from the projected vertices $v_i^s$ and faces $f_j$, (where $i\in N, j\in M$) via sampling, as explained in \cite{kato2018neural}. 
Different gradient flow rules are used to handle vertices residing outside or inside a given face. For ease of explanation, only the $x$ coordinate of a single vertex $v_i^s$ and a pixel $P_j$ are considered in each scenario. Note that the color of $P_j$ corresponds to a function $I_j(x_i)$, which is defined by freezing all variables except $x_i$ to compute the partial derivatives. 
As shown in Fig.~\ref{fig:rasterization}, during the rasterization process, the color $I_j(x_i)$ of a pixel $P_j$ changes instantly from $I(x_0)$ to $I_{ij}$ (i.e., starting and hitting point, respectively) when $x_i$ reaches a point $x_1$ where an edge of the face collides with $P_j$ center.
Note that such a change can be observed for pixels located either outside the face (Fig.~\ref{fig:rasterization}.i.b) or inside it (Fig.~\ref{fig:rasterization}.ii.b). 
If we then let $\delta_i^x=x_1-x_0$ be the distance traveled from a starting point $x_0$, and $\delta_j^I=I(x_1)-I(x_0)$ be the color change, it is straightforward to see that when computing the partial derivatives $\partial I_j(x_i)/\partial x_i$ during the rasterization process, they will be zero almost everywhere due to the sudden color change (Fig.~\ref{fig:rasterization}.i/ii.c). 
To address this issue, the authors of \cite{kato2018neural} introduce a gradual change between $x_0$ and $x_1$ via linear interpolation (Fig.~\ref{fig:rasterization}.i/ii.d), which allows for transformation of the partial derivates $\partial I_j(x_i)/\partial x_i$ into $\delta_j^I/\delta_i^x$ since the gradual change allows for non-zero derivatives (Fig.~\ref{fig:rasterization}.i/ii.e).
However, $I_j(x_i)$ has different left and right derivatives on $x_0$. Hence, the error signal $\delta_j^P$, which is back-propagated to pixel $P_j$ and indicates whether $P_j$ should be brighter or darker, is also used to handle the case where $x_i=x_0$. Finally, to deal with this situation, the authors of \cite{kato2018neural} define gradient flow rules to correctly modify pixels residing either outside or inside a given face during the backpropagation.

For $P_j$ residing outside the face, the gradient flow rule is defined as follows:
\begin{equation}
    \left. \frac{\partial I_j(x_i)}{\partial x_i}\right\rvert_{x_i=x_0} = \begin{cases} \frac{\delta_j^I}{\delta_i^x}, & \mbox{if } \delta_j^P\delta_j^I<0; \\ 0, & \mbox{if } \delta_j^P\delta_j^I\ge0, \end{cases}
    \label{eq:pj_out}
\end{equation}
where $\delta_i^x=x_1-x_0$ indicates the distance traveled by $x_i$ during the rasterization procedure; $\delta_j^I=I(x_1)-I(x_0)$ represents the color change; and $\delta_j^P$ corresponds to the error signal backpropagated to pixel $P_j$. 
As described in \cite{kato2018neural}, to minimize the loss during training, pixel $P_j$ must become darker for $\delta_j^P>0$. In fact, since the sign of $\delta_j^I$ denotes whether $P_j$ can be brighter or darker, for $\delta_j^I>0$, pixel $P_j$ becomes brighter when pulling $x_i$ toward the face but, at the same time, $P_j$ cannot become darker when moving $x_i$ as it would require a negative $\delta_j^I$. Thus, the gradient should not flow if $\delta_j^P>0\:\wedge\:\delta_j^I>0$.
In addition, the face and $P_j$ might also not overlap, regardless of where $x_i$ moves, as the hitting point $x_1$ may not exist.
In this case, the derivatives are defined as $\partial I_j(x_i)/\partial x_i\rvert_{x_i=x_0}=0$. 
This means that the gradient should never flow when $\delta_j^P\delta_j^I\ge0$, in accordance with Eq.~\ref{eq:pj_out}.
Finally, we note that the derivative with respect to $y_i$ is obtained by using the y-axis in Eq.~\ref{eq:pj_out}.

For a point $P_j$ residing inside the face, the left and right derivatives are first defined \cite{kato2018neural} at $x_0$ as $\delta_j^{Ia}=I(x_1^a)-I(x_0)$, $\delta_j^{Ib}=I(x_1^b)-I(x_0)$, $\delta_x^a=x_1^a-x_0$, and 
$\delta_x^b=x_1^b-x_0$. Then, in a similar way to a point $P_j$ residing outside the face, they define the gradient flow rules as follows:
\begin{equation}
    \left. \frac{\partial I_j(x_i)}{\partial x_i}\right\rvert_{x_i=x_0} = \left. \frac{\partial I_j(x_i)}{\partial x_i}\right\rvert_{x_i=x_0}^a + \left. \frac{\partial I_j(x_i)}{\partial x_i}\right\rvert_{x_i=x_0}^b,
\end{equation}
\begin{equation}
    \left. \frac{\partial I_j(x_i)}{\partial x_i}\right\rvert_{x_i=x_0}^a = \begin{cases} \frac{\delta_j^{I^a}}{\delta_x^a}, & \mbox{if } \delta_j^P\delta_j^{I^a}<0; \\ 0, & \mbox{if } \delta_j^P\delta_j^{I^a}\ge0, \end{cases}
\end{equation}
\begin{equation}
    \left. \frac{\partial I_j(x_i)}{\partial x_i}\right\rvert_{x_i=x_0}^b = \begin{cases} \frac{\delta_j^{I^b}}{\delta_x^b}, & \mbox{if } \delta_j^P\delta_j^{I^b}<0; \\ 0, & \mbox{if } \delta_j^P\delta_j^{I^b}\ge0. \end{cases}
\end{equation}

\subsection{Multi-task Loss}\label{subsec:multi_task_loss}

The final outputs of the proposed framework are the 3D joint positions and the hand mesh. However, in real-world datasets, the ground truths for 3D hand mesh, pose, shape, and view parameters, which are unknown in unconstrained situations, are hard to collect. Our framework therefore automatically generates intermediate representations (i.e., 2D heatmaps, hand silhouettes, and 2D re-projected joint positions and meshes), which are then exploited to train the whole system jointly using a single loss function. 
The ground truths in this case are defined as follows:
\begin{itemize}
    \item 2D heatmaps (one per joint) are built using a 2D Gaussian with a standard deviation of $2.5$ pixels, centered on the 2D joint locations annotations identified in the datasets, to describe the likelihood of a given joint residing in that specific area;
    \item Hand silhouettes are computed from the input image with the GrabCut algorithm, implemented using the OpenCV library, and 2D joint annotations from the datasets are used to initialize the foreground, background, and probable foreground/background regions, following \cite{boukhayma20193d};
    \item The 3D joint positions and 2D re-projected joint positions are compared directly with the 3D and 2D joint positions provided by the various datasets;
    \item The 2D re-projected meshes are compared with the same hand silhouette masks built with the GrabCut algorithm.
\end{itemize}
Formally, the multi-task loss function employed here is defined through the following equation:
\begin{equation}
    L = L_{SFE} + L_{3D} + L_{2D} + L_{M} + L_{Reg},
\end{equation}
where $L_{SFE}$, $L_{3D}$, $L_{2D}$, $L_{M}$, and $L_{Reg}$, represent the semantic feature extractor (i.e., 2D heatmaps and hand silhouette), 3D joint positions, 2D joint re-projection, hand silhouette mask (i.e., the re-projected mesh), and model parameter regularization losses, respectively. 

\subsubsection{$L_{SFE}$} 
The semantic feature extractor estimates 2D heatmaps and the hand silhouette. The loss for a given hourglass module $h_i$ is defined as the sum of heatmaps $L_2$ and the pixel-wise binary cross-entropy (BCE) losses for the silhouette, as follows:
\begin{equation}
    L_{h_i} = \lVert H - \hat{H} \rVert_2^2 + \lVert M - \hat{M} \rVert_{BCE}^2,
\end{equation}
where $\hat{\cdot}$ is the hourglass output for the 2D heatmaps $H$ and the silhouette mask $M$, for which the ground truths are derived via the 2D Gaussian and GrabCut algorithm, respectively.
The two stacked hourglass network losses are then summed to apply intermediate supervision since, as demonstrated in \cite{newell2016stacked}, this improves the final estimates. Thus, the SFE loss is defined in the following equation:
\begin{equation}
    L_{SFE} = L_{h_1} + L_{h_2},
\end{equation}

\subsubsection{$L_{3D}$} 
An $L_2$ loss is also used to measure the distance between the estimated 3D joint positions $RJ(\beta,\theta)$ and the ground truth coordinates $w_{3D}$ provided by the datasets, as follows:
\begin{equation}
    L_{3D} = \lVert w_{3D} - RJ(\beta,\theta) \rVert_2^2
\end{equation}

\subsubsection{$L_{2D}$} 
The 2D re-projected hand joint positions loss is used to refine the view parameters $t$, $s$, and $R$, for which the ground truths are generally unknown. It is computed as follows:
\begin{equation}
    L_{2D} = \lVert w_{2D} - \hat{w}_{2D} \rVert_1,
\end{equation}
where $\hat{w}_{2D}$ indicates the network 2D re-projected positions, and $w_{2D}$ are the ground truths of 2D joint positions annotated in a given dataset.
Notice that an $L_1$ loss is used since it is less sensitive and more robust to outliers than the $L_2$ loss. 

\subsubsection{$L_{M}$} 
The silhouette mask $M$ loss is introduced into the weak supervision, since the hand mesh should be consistent with its silhouette \citep{zhang2019end} or depth map \citep{kato2018neural}. This $L_2$ loss therefore helps to refine both the hand shape and the camera view parameters, via the following equation:
\begin{equation}
    L_M = \lVert M - \hat{M} \rVert_2^2,
\end{equation}
where $\hat{M}$ is the 2D re-projected mesh, and $M$ corresponds to the hand silhouette mask used as ground truth, extracted using the GrabCut algorithm.

\subsubsection{$L_{Reg}$} 
The last loss component is a regularization term that is applied with the aim of reducing the magnitudes of the hand model parameters $\beta$ and $\theta$, in order to avoid unrealistic mesh representations. Focusing only on the 2D and 3D joint positions while ignoring the hand surfaces results in the mesh fitting joint locations but completely ignoring the actual anatomy of the hand. Hence, to avoid possible extreme mesh deformations, a regularization term is used as follows:
\begin{equation}
    L_{Reg} = \lVert \beta \rVert_2^2 + \lVert \theta \rVert_2^2.
\end{equation}

\section{Experimental Results}\label{sec:results}
In this section, we first introduce the benchmark datasets for 3D hand pose and shape estimation and hand gesture recognition that are used to validate our framework, and the data augmentation strategy that is employed to better exploit all the available samples. 
We then present a comprehensive performance evaluation. We report the results of ablation studies on each component of the framework to highlight the effectiveness of the proposed approach both quantitatively and qualitatively. We conduct a comparison with state-of-the-art alternatives for the 3D hand pose and shape estimation, and present the results obtained from a different task (i.e., hand-gesture recognition), so that the abstraction capabilities of our framework can be fully appreciated.

\subsection{Datasets}
The following benchmark datasets are exploited to evaluate our framework: the synthetic object manipulation (ObMan) \citep{hasson2019learning}, stereo hand dataset (STB) \citep{zhang20163d},  Rhenish-Westphalian Technical University gesture (RWTH) \citep{dreuw2006modeling}, and the creative Senz3D (Senz3D) \citep{minto2015exploiting} datasets. Specifically, ObMan is used to pre-train the SFE to generate 2D heatmaps and hand silhouette estimations that are as accurate as possible; STB is employed to evaluate the 3D hand pose and shape estimations through ablation studies and comparisons with state-of-the-art methods; and RWTH and Senz3D are utilized to assess the generalization capabilities of our framework to the task of hand gesture recognition.

\subsubsection{ObMan} 
This is a large-scale synthetic dataset containing images of hands grasping different objects such as bottles, bowls, cans, jars, knives, cellphones, cameras, and remote controls. Realistic images of embodied hands are built by transferring different poses to hands via the SMPL+H model \citep{romero2017embodied}. Several rotation and translation operations are applied to maximize the viewpoint variability to provide natural occlusions and coherent backgrounds. For each hand-object configuration, object-only, hand-only, and hand-object images are generated with the corresponding segmentation, depth map, and 2D/3D joints location of 21 keypoints. From this dataset, we selected 141,550 RGB images with dimensions $256\times256$, showing either hand-only or hand-object configurations, to train the semantic feature extractor.

\subsubsection{STB} 
This dataset contains stereo image pairs (STB-BB) and depth images (STB-SK), and was created for the evaluation of hand pose tracking/estimation difficulties in real-world scenarios. 
Twelve different sequences of hand poses were collected with six different backgrounds representing static or dynamic scenes.
The hand and fingers are either moved slowly or randomly to give both simple and complex self-occlusions and global rotations. 
Images in both collections have the same resolution of $640\times480$, identical camera pose, and similar viewpoints. 
Furthermore, both subsets contain 2D/3D joint locations of 21 keypoints. 
From this collection, we used only the STB-SK subset to evaluate the proposed network, and divided it into 15,000 and 3,000 samples for the training and test sets, respectively.

\subsubsection{RWTH} 
This dataset includes fingerspelling gestures from the German sign language. It consists of RGB video sequences for 35 signs representing letters from A to Z, the 'SCH' character, umlauts \"A, \"O and \"U, and numbers from one to five. For each gesture, 20 different individuals were recorded twice, using two distinct cameras with different viewpoints, at resolutions of $320\times240$ and $352\times288$, giving a total of 1,400 samples. From this collection, we excluded all gestures requiring motion, i.e., the letters J, Z, \"A, \"O and \"U. The final subset contained 30 static gestures over 1,160 images. This collection was divided into disjoint training and test sets containing 928 and 232 images, respectively, in accordance with \citep{zimmermann2017learning}.

\subsubsection{Senz3D}
This dataset contains 11 different gestures performed by four different individuals. To increase the complexity of the samples, the authors collected gestures with similar characteristics (e.g., the same number of raised fingers, low distances between fingers, and touching fingertips). All gestures were captured using an RGB camera and a time-of-flight (ToF) depth sensor at a resolution of $320\times240$. Moreover, each gesture was repeated by each person 30 times, for a total of 1,320 acquisitions. All of the available samples from this collection were employed in the experiments.

\subsection{Data Augmentation}
Data augmentation is a common practice that can help a model to generalize the input data, making it more robust. In this work, up to four different groups of transformations, randomly selected during each iteration of the training phase, were applied to an input image to further increase the dissimilarities between samples. These transformations were as follows:

\begin{itemize}
    \item \textit{blur:} This is obtained by applying a Gaussian filter with varying strength, via $\sigma \in [1,3]$ kernel, or by computing the mean over a neighborhood using a kernel shape with a size of between $3\times3$ and $9\times9$;
    \item \textit{random noise:} This is achieved by adding Gaussian noise to an image, either sampled randomly per pixel channel or once per pixel from a normal distribution $\mathcal{N}(0,0.05\cdot255)$;
    \item \textit{artificial occlusion:} This can be attained either by dropping (i.e., setting to black) up to 30\% of the contiguous pixels, or by replacing up to 20\% pixels using a salt-and-pepper strategy;
    \item \textit{photometric adjustments:} These are derived from arithmetic operations applied to the image matrix, for example by adding a value in the range $[-20, 20]$ to each pixel, by improving or worsening the image contrast, or by changing the brightness by multiplying the image matrix with a value in the range $[0.5, 1.5]$.
\end{itemize}
Note that all transformations only affect the appearance of an image, and leave the 2D/3D coordinates unaltered.

\subsection{Performance Evaluation}
The proposed system was developed using the Pytorch framework. All experiments were performed using an Intel Core i9-9900K @3.60GHz CPU, 32GB of RAM, and an Nvidia GTX 1080 GPU with 8GB GDDR5X RAM.
With this configuration, at inference time, the SFE, VE, and HE components required $64.9$, $9.6$, and $15.04$ ms, respectively, giving a total of $89.54$ ms per input image. The proposed system can therefore analyze approximately 11 images per second, regardless of the dataset used. Moreover, we note that this speed could be further improved with higher performance hardware. In fact, when using a more recent GPU model such as the Nvidia GeForce RTX 3080 with 10GB GDDR6X RAM, the total time required to analyze a single input image is reduced to $37.02$ ms, enabling about 27 images to be examined per second, i.e., a speed increase of roughly 2.4x with respect to our configuration.

To evaluate the proposed framework, we employ three metrics that are commonly used for 3D hand pose and shape estimation, and for hand gesture recognition. These are the 3D end-point-error (EPE) and area under the curve (AUC) for the former task, and accuracy for the latter. 
The EPE used in the ablation studies is defined as the average Euclidean distance, measured in millimeters (mm), between predicted and ground truth keypoints. The AUC is computed based on the percentage of correct 3D keypoints (3D PCK) at different thresholds, for a range of 20-50 mm. Finally, for both metrics, the public implementation in \cite{zimmermann2017learning} is employed for a fair comparison. 

\begin{table}[t]
	\centering
	\caption{Semantic feature extractor ablation study.}
	\label{tab:sfe_ablation}
		\begin{tabular}{p{.8\columnwidth} c}
        	\hline
        	\textbf{Design choice} & \textbf{EPE} \\
        	\hline
        	No semantic feature extractor (SFE) & 13.06 \\
        	SFE (heatmaps only) & 11.69 \\
        	SFE (heatmaps + silhouette) & 11.52 \\
        	ObMan pre-trained SFE (heatmaps + silhoutte) & 11.12 \\
        	\hline
		\end{tabular}
\end{table}

\begin{table}[t]
	\centering
	\caption{Viewpoint encoder ablation study.}
	\label{tab:ve_ablation}
		\begin{tabular}{p{.75\columnwidth} c}
        	\hline
        	\textbf{Design choice} & \textbf{EPE} \\
        	\hline
        	2*ResNet modules VE & 28.77 \\
        	4*ResNet modules VE & 11.12 \\
        	5*ResNet modules VE & 12.25 \\
        	\hline
        	3*1024 dense layers VE (VE$_1$) & 10.79 \\ 
        	2*2048/1*1024 dense layers VE (VE$_2$) & 11.12 \\ 
        	3*2048 dense layers VE & 11.86 \\ 
        	\hline
		\end{tabular}
\end{table}

\begin{table}[t]
	\centering
	\caption{Hand pose/shape estimator ablation study.}
	\label{tab:hpse_ablation}
		\begin{tabular}{p{.75\columnwidth} c}
        	\hline
        	\textbf{Design choice} & \textbf{EPE} \\
        	\hline
        	15 PCA parameters MANO layer & 12.31 \\
        	30 PCA parameters MANO layer & 11.47 \\
        	45 PCA parameters MANO layer & 11.12 \\
			\hline
			No 2D re-projection & 11.56 \\
			With 2D re-projection & 11.12 \\
			\hline
			No $L_{Reg}$ loss & 10.10 \\
			With $L_{Reg}$ loss & 11.12 \\
			\hline
		\end{tabular}
\end{table}

\begin{table}[t]
	\centering
	\caption{Advanced components ablation study.}
	\label{tab:adv_ablation}
		\begin{tabular}{p{.75\columnwidth} c}
        	\hline
        	\textbf{Design choice} & \textbf{EPE} \\
        	\hline
			No adapt skeleton \& hourglass summation & 11.12 \\
			With adapt skeleton \& hourglass summation & 9.10 \\
			With adapt skeleton \& hourglass concatenation \& (VE$_1$) & 9.03 \\
			With adapt skeleton \& hourglass concatenation \& (VE$_2$) & 8.79 \\
			\hline
		\end{tabular}
\end{table}

\subsubsection{Framework Quantitative and Qualitative Results}\label{subsubsec:framework_quantitative_and_qualitative}
The proposed framework contains several design choices that were made in order to obtain stable 3D hand pose and shape estimations. 
We therefore performed ablation studies to assess the effectiveness of each of these decisions. The obtained results are summarized in Tables \ref{tab:sfe_ablation}, \ref{tab:ve_ablation}, \ref{tab:hpse_ablation}, and \ref{tab:adv_ablation}, where each table shows the results for a given component, i.e., the SFE, VE, HE, and advanced framework components.
All of the reported EPE scores were computed for the STB dataset, while pre-training of the SFE unit was carried out exclusively on the ObMan dataset, since it contains a high number of synthetic images under various conditions. 
For both collections, mini-batches of size six and an Adam optimizer with a learning rate of $10^{-4}$ and a weight decay of $10^{-5}$ were used to train the system. The framework was trained for 60 and 80 epochs on the ObMan and STB datasets, respectively, as the former contained substantially more samples.
In relation to the STB training time, which involves the entire framework, each mini-batch required $\sim$0.5 seconds to be analyzed by the specified hardware configuration, giving a total of $\sim$1278 s per training epoch. 

The first experiment quantitatively evaluated the usefulness of the SFE component in terms of producing more accurate 3D hand pose and shape estimations. As shown in Table~\ref{tab:sfe_ablation}, while it is still possible to achieve estimations by feeding the input image directly to the VE (i.e., with no SFE), an improvement of $\sim$1.5 mm in the estimation can be obtained by extracting the semantic features. This indicates that the 2D heatmaps allow the network to focus on the positions of the hand joints, whereas generating the heatmaps and silhouette simultaneously enables a comprehensive view of the hand that forces the joints to be placed in the right position when moving to a 3D plane. This behavior is further supported by the results of pre-training the SFE component on the ObMan dataset, where the occlusions force the network to create meaningful abstractions for both the 2D heatmaps and silhouette. 

The second test gauged the quality of the VE in terms of estimating view parameters $v$. As shown in Table~\ref{tab:ve_ablation}, using either a low or high number of ResNet modules (i.e., two in the former case or five or more in the latter) to produce the latent space representation $l_v$ results in an increased EPE score, which is often associated with underfitting and overfitting. Slightly better performances can be achieved by reducing the sizes of the dense layers used to build up the vector $v$. However, although the smaller VE (i.e., $VE_1$) can perform better than a larger one (i.e., $VE_2$), this result does not apply when extra steps are included, such as skeleton adaptation and hourglass output concatenation (shown in Table~\ref{tab:adv_ablation}), suggesting that some information can still be lost.

\begin{figure*}[t]
	\begin{overpic}[width=\textwidth]{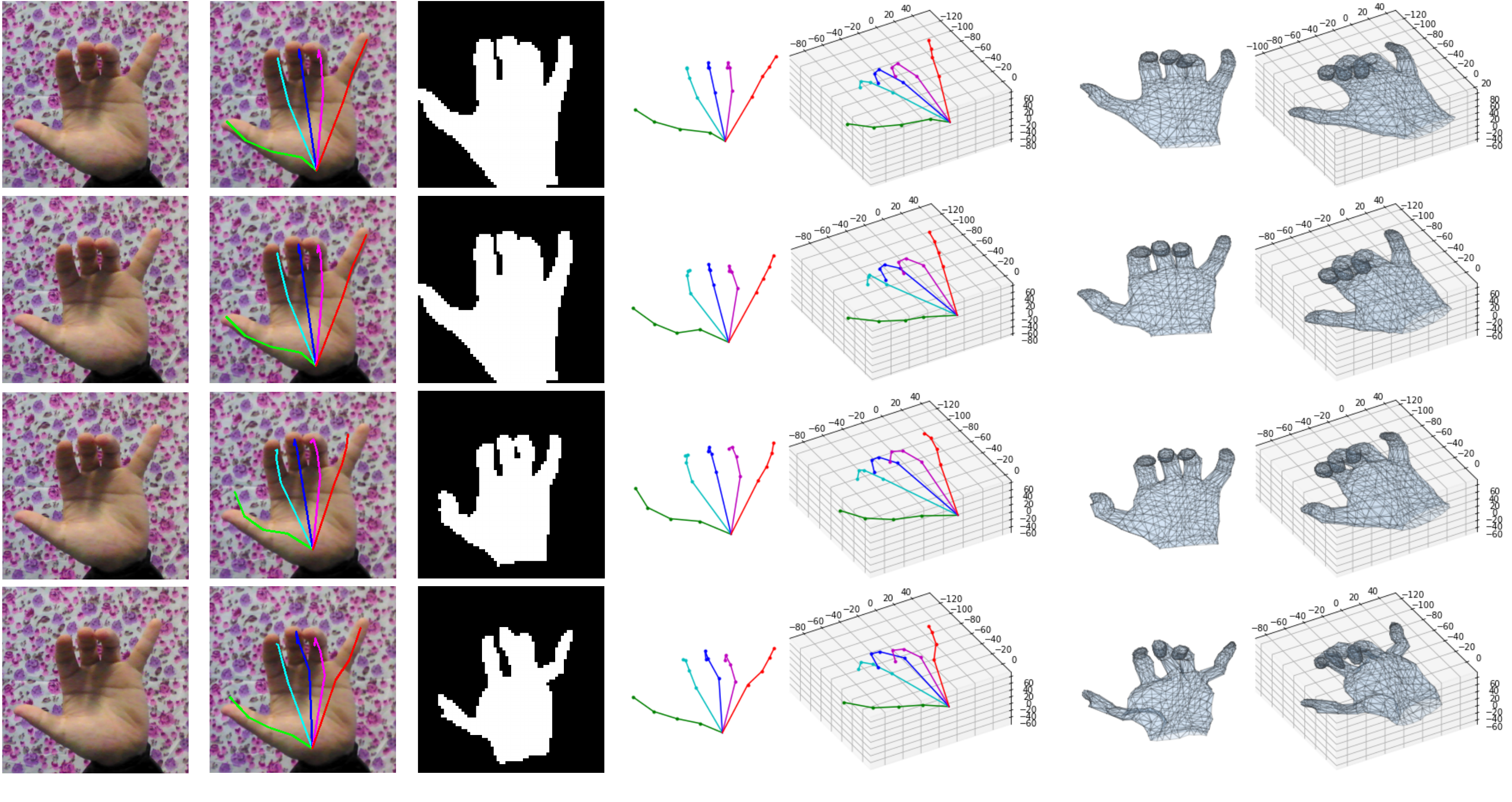}
		\put(5.6,0){(a)}	
		\put(19.1,0){(b)}	
		\put(32.9,0){(c)}	
		\put(51.2,0){(d)}	
		\put(82.2,0){(e)}	
	\end{overpic}
	\caption{STB dataset 3D pose and shape estimation outputs. From top to bottom, the presented framework, framework without 2D re-projection, framework without SFE module, and full framework without regularization loss. Input image, 2D joints, silhouette, 3D joints, and mesh, are shown in (a), (b), (c), (d), and (e), respectively.}
	\label{fig:ablation_qualitative}
\end{figure*}

The third experiment, which is summarized in Table~\ref{tab:hpse_ablation}, focused on the hand pose and shape estimator. Tests were performed on the number of parameters of the MANO layer, the proposed 2D re-projection, and the regularization loss. Increasing the number of hand articulations allowed us, as expected, to obtain more realistic hands and consequently more precise estimations when all 45 values were used. Applying the proposed 2D re-projection further reduced the EPE score by providing the MANO layer with direct feedback on its output. Employing the regularization loss resulted in a higher keypoint distance, with the difference derived from the hand shape collapsing onto itself as shown in the bottom row of Fig.~\ref{fig:ablation_qualitative}.

The fourth and last test, reported in Table~\ref{tab:adv_ablation}, dealt with advanced strategies, i.e., skeleton adaptation (adapt skeleton) and hourglass output concatenation, rather than summation. The former strategy allowed for a significant performance boost (a 2 mm lower EPE) since it directly refines the 3D joints produced by the MANO layer. Stacked hourglass output concatenation improved the precision of the system by a further 0.31 mm, since it provided the VE with a finegrained input representation. However, this detailed input description requires bigger dense layers (i.e., $VE_2$) to avoid losing any information. Consequently, using smaller dense layers (i.e., $VE_1$) results in an increase in the EPE score.

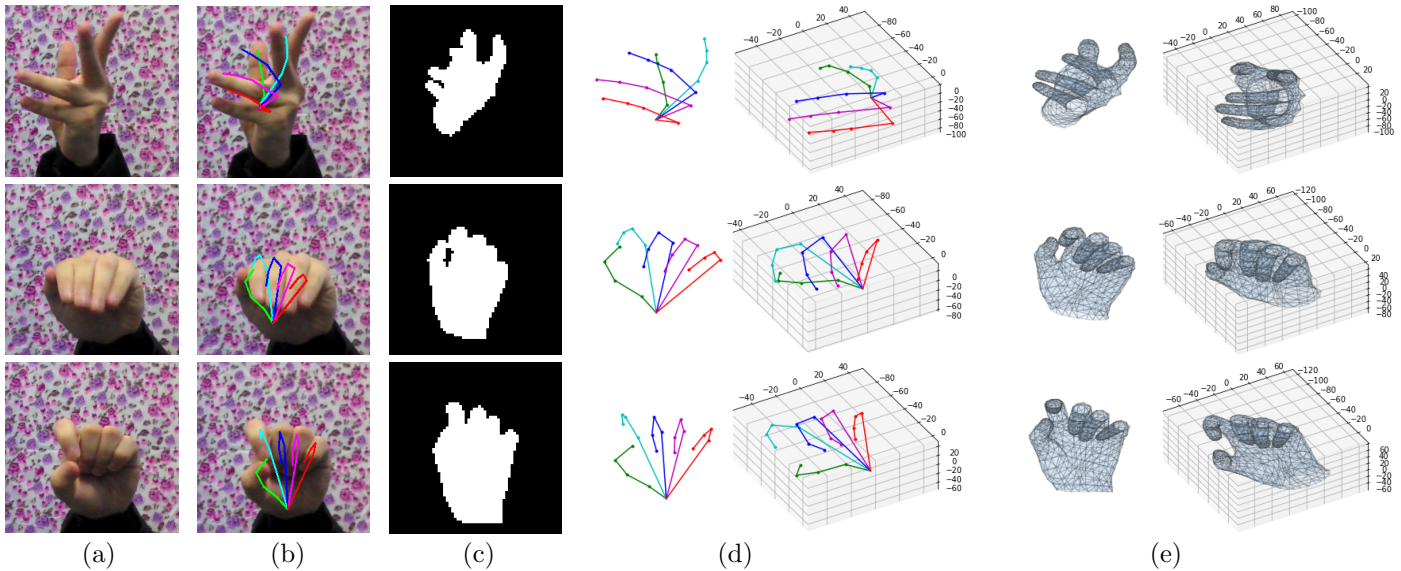
\begin{figure*}[t]
	\begin{overpic}[width=\textwidth]{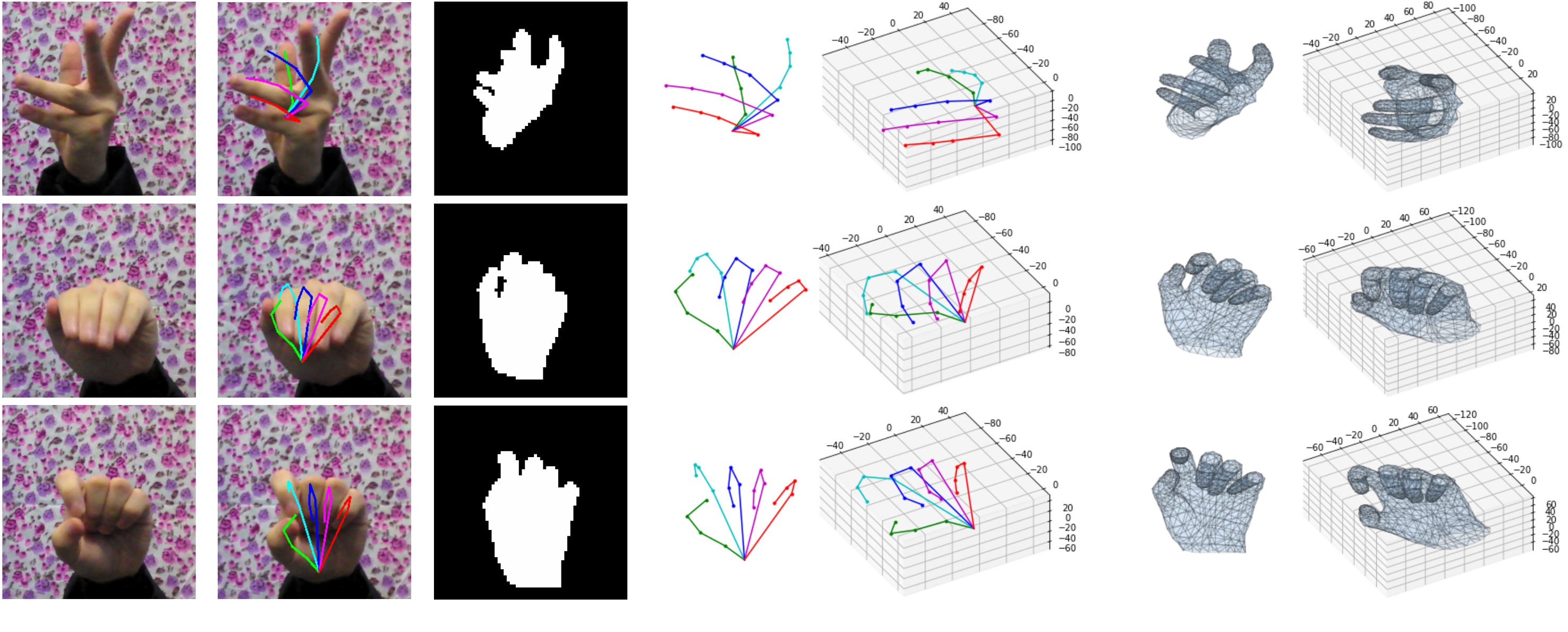}
		\put(5.6,0){(a)}	
		\put(19.1,0){(b)}	
		\put(32.9,0){(c)}	
		\put(51.2,0){(d)}	
		\put(82.2,0){(e)}	
	\end{overpic}
	\caption{STB dataset failed 3D pose and shape estimation outputs for the presented framework. Input image, 2D joints, silhouette, 3D joints, and mesh, are shown in (a), (b), (c), (d), and (e), respectively.}
	\label{fig:failure_qualitative}
\end{figure*}

\begin{table*}[t]
	\centering
	\caption{AUC state-of-the-art comparison on STB dataset. Works are subdivided according to their input and output types.}
	\label{tab:auc_comp}
		\begin{tabular}{l ccc}
        	\hline
        	\textbf{Model} & \textbf{Input} & \textbf{Output} & \textbf{AUC} \\
        	\hline
        	CHPR \citep{sun2015cascaded} & Depth & 3D skeleton & 0.839 \\
        	ICPPSO \citep{qian2014realtime} & RGB-D & 3D skeleton & 0.748 \\
        	PSO \citep{oikonomidis2011efficient} & RGB-D & 3D skeleton & 0.709 \\
        	Dibra et al. \cite{dibra2018monocular} & RGB-D & 3D skeleton & 0.923 \\
        	\hline
        	Cai et al. \cite{cai20203d} & RGB & 3D skeleton & 0.996 \\
        	Iqbal et al. \cite{iqbal2018hand} & RGB & 3D skeleton & 0.994 \\
        	Hasson et al. \cite{hasson2019learning} & RGB & 3D skeleton & 0.992 \\
        	Yang and Yao \cite{yang2019disentangling} & RGB & 3D skeleton & 0.991 \\
        	Spurr et al. \cite{spurr2018cross} & RGB & 3D skeleton & 0.983 \\
        	Zummermann and Brox \cite{zimmermann2017learning} & RGB & 3D skeleton & 0.986 \\
        	Mueller et al. \cite{mueller2018ganerated} & RGB & 3D skeleton & 0.965 \\
        	Panteleris et al. \cite{panteleris2018using} & RGB & 3D skeleton & 0.941 \\
        	\hline
        	Ge et al. \cite{ge20193d} & RGB & 3D skeleton+mesh & 0.998 \\
        	Baek et al. \cite{baek2020weakly} & RGB & 3D skeleton+mesh & 0.995 \\ 
        	Zhang et al. \cite{zhang2019end} & RGB & 3D skeleton+mesh & 0.995 \\
        	Boukhayma et al. \cite{boukhayma20193d} & RGB & 3D skeleton+mesh & 0.993 \\
			ours & RGB & 3D skeleton+mesh & 0.995 \\
			\hline
		\end{tabular}
\end{table*}

The differences in the the qualitative results for different framework configurations are shown in Fig.~\ref{fig:ablation_qualitative}. From the top row to the bottom, the outputs correspond to the proposed framework, the framework without 2D re-projection, the framework without the SFE module, and the full framework without the regularization loss. As can be seen, the most important component in terms of obtaining coherent hand shapes is the regularization loss, since otherwise the mesh collapses onto itself in order to satisfy the 3D joint locations during training time (bottom row in Fig.~\ref{fig:ablation_qualitative}.c and Fig.~\ref{fig:ablation_qualitative}.e). 
When we employ the SFE module (first two rows in Fig.~\ref{fig:ablation_qualitative}), more accurate 3D joints and shapes are generated since the SFE enforces both the correct localization of joints and the generation of a more realistic silhouette (Figs.~\ref{fig:ablation_qualitative}.b, c, and d). 
When we re-project the generated 3D-coordinates and mesh, the final 3D joint locations and hand shape (Figs.~\ref{fig:ablation_qualitative}.d and e) are more consistent with both the estimated 2D locations and the input image (Figs.~\ref{fig:ablation_qualitative}.b and a). To conclude this qualitative evaluation, some examples of failed 3D pose and shape estimations are shown in Fig.~\ref{fig:failure_qualitative}. It can be seen that although coherent hand poses and shapes are generated, the framework is unable to produce the correct output due to wrong estimations of both the 2D joints and the silhouette by the SFE. This preliminary error is then amplified by the subsequent framework modules, as can be seen from the discrepancy between the 2D and 3D joint locations shown in Figs.~\ref{fig:failure_qualitative}.b and d. Although the loss described in Section~\ref{subsec:multi_task_loss} ultimately forces the MANO layer to produce consistent hands, as also discussed for the third experiment, it can also result in greater inaccuracy in the 3D joint position estimation to guarantee such a consistency. This outcome has two implications: firstly, it indicates that there is still room for improvement, particularly for the 2D joints and silhouette estimations that represent the first step in the proposed pipeline; and secondly, it highlights the effectiveness of the proposed framework, which can generate stable 3D representations from RGB images.

\subsubsection{Comparison of 3D Hand Pose/Shape Estimation} 
To demonstrate the effectiveness of the proposed framework, a state-of-the-art 3D PCK AUC comparison was carried out, as shown in Table~\ref{tab:auc_comp}. As can be seen, the presented system is competitive with other successful schemes while using only RGB images and outputting both 3D skeleton locations and mesh, indicating that all of our design choices allow the framework to generate good estimates using only RGB information. It is particularly interesting that the proposed method was able to easily outperform systems that exploited depth data, suggesting that the simultaneous use of the multi-task SFE, VE, and 2D re-projection can help to produce correct estimations by compensating for the missing depth information. 
Specifically, the multi-task SFE enables the implementation of a customized VE that, unlike the scheme in \cite{boukhayma20193d}, does not require to be pre-trained on a synthetic dataset in order to perform well; there is also no need to normalize the latent feature space representation by using a VAE to disentangle different factors influencing the hand representation in order to obtain accurate 3D hand poses, as described in \cite{yang2019disentangling}. 
Furthermore, thanks to the re-projection module, these results are obtained without applying an iterative regression module to the MANO layer, unlike in \cite{zhang2019end}, and \cite{baek2020weakly}, where progressive changes are carried out recurrently to refine the estimation parameters, thus simplifying the training procedure. In addition, the solutions implemented in the proposed framework allow us to avoid input assumptions and post-processing operations, unlike the majority of schemes in the literature where some parameter (e.g., global hand scale or root joint depth) is assumed to be known at test time, and secondly to achieve similar performance to the best model devised in \cite{ge20193d}, even though the latter scheme employs a more powerful solution (i.e., a graph CNN instead of a fixed MANO layer) for the 3D hand pose and shape estimation.

\begin{figure}[t]
    \centering
    \includegraphics[width=\columnwidth]{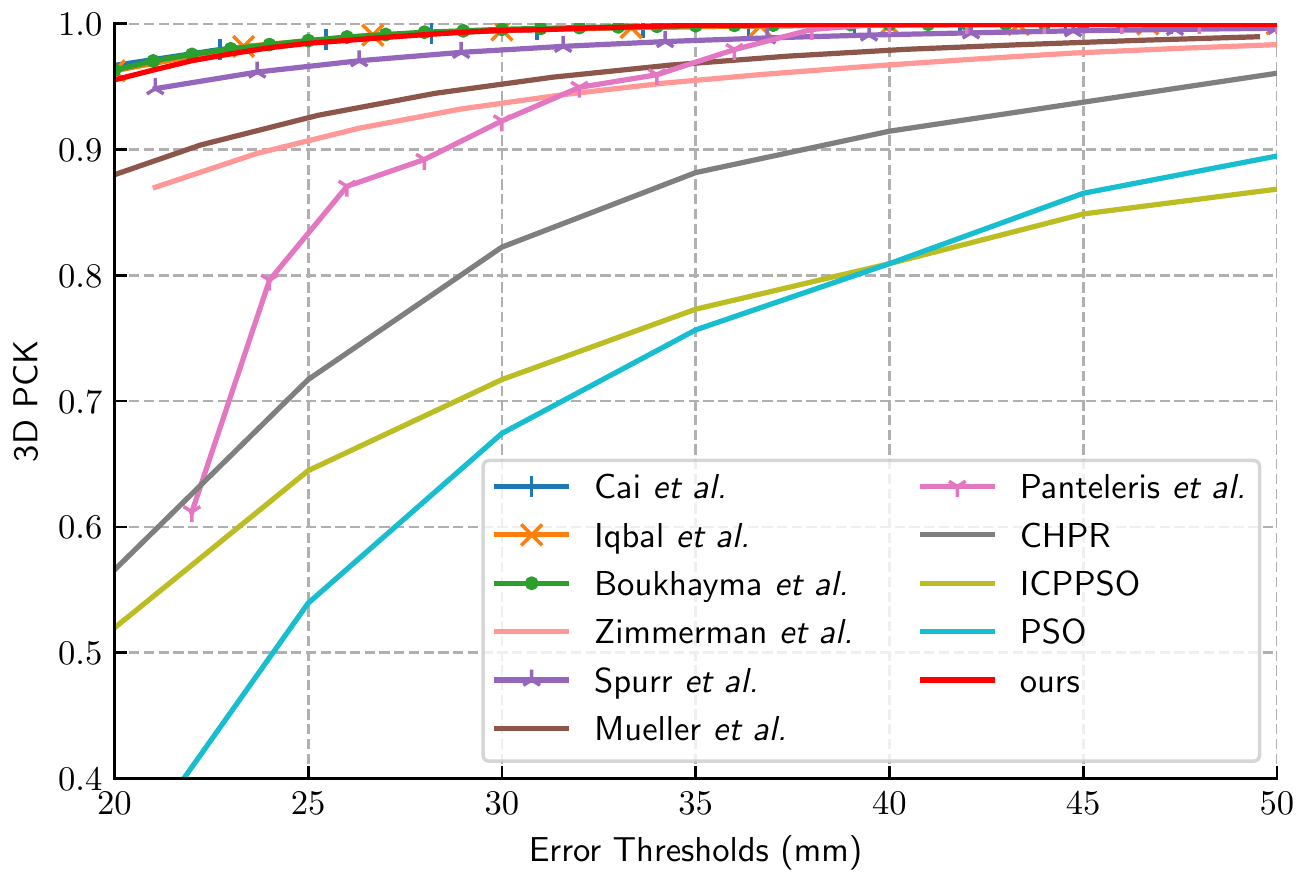}
    \caption{3D PCK state-of-the-art comparison on STB dataset.}
    \label{fig:3d_pck}
\end{figure}

To conclude this comparison, the 3D PCK curve computed at different thresholds for several state-of-the-art works is shown in Fig.~\ref{fig:3d_pck}. It can be seen that performance on the 3D hand pose and shape estimation task is becoming saturated, and newer works can consistently achieve high performance with low error thresholds. In this context, the proposed method is on par with the top works in the literature, further supporting the view that all of our design choices allowed the framework to generate good 3D poses and shapes from monocular RGB images.

\subsubsection{Comparison on Hand Gesture Recognition} 
To assess the generalizability of the proposed framework, experiments were performed on the RWTH and Senz3D datasets. Since our architecture does not include a classification component, it was extended by attaching the same classifier described in \cite{zimmermann2017learning} to handle the new task. This classifier takes as input the 3D joint coordinates generated by the MANO layer, and consists of three fully connected layers with a ReLU activation function.
Note that all of the weights (except those used for the classifier) are frozen when training the system on the hand gesture recognition task, so that it is possible to correctly evaluate the generalizability of the framework. 
Moreover, although weights are frozen, the entire framework still needs to be executed. Hence, the majority of the training time is spent on the generation of the 3D joint coordinates, since this is where most of the computation is performed; as a result, each mini-batch is analyzed by the specified hardware configuration in $\sim$222 ms (i.e., $\sim$37 ms per image), giving total times of $\sim$206 s and $\sim$235 s per epoch for the RWTH and Senz3D datasets, respectively.
Our experiments followed the testing protocol devised in \cite{dibra2018monocular} in order to present a fair comparison; this consisted of 10-fold cross-validation with non-overlapping 80/20 splits for the training and test sets, respectively. In a similar way to other state-of-the-art works, all images were cropped close to the hand to remove as much background as possible and meet the requirements for the input size, i.e., $256\times256$.
The results are shown in Table~\ref{tab:gesture_recog_comp}, and a comparison with other schemes in the literature is presented.
It can be seen that our framework consistently outperformed another work focusing on 3D pose and shape estimation (i.e., \cite{zimmermann2017learning}) on both datasets, meaning that it generates more accurate joint coordinates from the RGB image; the same result was also obtained for the estimation task, as shown in Table~\ref{tab:auc_comp}.
However, methods that exploit depth information (i.e., \cite{dibra2018monocular} or concentrate on hand gesture classification (i.e., \cite{papadimitriou2019fingerspelled}) can still achieve slightly higher performance. There are two reasons for this. 
Firstly, by concentrating on the hand gesture classification task, lower performance is achieved on the estimation task, although similar information, such as the 3D joint locations, is used. As a matter of fact, even though they exploit depth information in their work, the authors of \cite{dibra2018monocular} obtained an AUC score of 0.923, while the scheme in \cite{zimmermann2017learning} and the proposed framework achieved AUC scores of 0.986 and 0.994 on the STB dataset for 3D hand pose estimation, respectively. Secondly, as discussed in Section~\ref{subsubsec:framework_quantitative_and_qualitative} and shown by the qualitative results in Fig.~\ref{fig:failure_qualitative}, the proposed architecture could be improved further by increasing the estimation accuracy of the 2D joints and silhouette, indicating that if a good hand abstraction is used to derive the 3D hand pose and shape, this can be effective for the hand gesture recognition task.

In summary, the proposed method achieves state-of-the-art performance on the 3D hand pose and shape estimation task, can outperform other existing estimation approaches when applied to the hand gesture recognition task, and behaves in a comparable way to other specifically designed hand gesture recognition systems. This indicates that the proposed pipeline outputs stable hand pose estimations that can be effectively used to recognize hand-gestures.

\begin{table}[t]
	\centering
	\caption{Hand-gesture recognition accuracy comparison.}
	\label{tab:gesture_recog_comp}
		\begin{tabular}{l cc}
        	\hline
        	\textbf{Model} & \textbf{RWTH} & \textbf{Senz3D} \\
        	\hline
        	Papadimitriou and Potamianos \cite{papadimitriou2019fingerspelled} & 73.92\% & - \\
        	Memo and Zanuttigh \cite{memo2018head} & - & 90.00\% \\
        	Dibra et al. \cite{dibra2018monocular} & 73.60\% & 94.00\% \\
        	Dreuw et al. \cite{dreuw2006modeling} & 63.44\% & - \\
        	Zimmerman and Brox \cite{zimmermann2017learning}* & 66.80\% & 77.00\% \\
        	ours* & 72.03\% & 92.83\% \\
			\hline
		\end{tabular}\\
	\footnotesize{$^*$method focusing on 3D hand pose and shape estimation}
\end{table}

\section{Conclusion}\label{sec:conclusions}
In this paper, we have presented an end-to-end framework for the estimation of 3D hand pose and shape, and successfully applied it to the task of hand gesture recognition. 
Our model comprises three modules, a multi-task SFE, a VE, and an HE with weak re-projection, each of which has certain strengths and weaknesses. 
More specifically, the SFE, thanks to its pre-training providing multi-task context, enables our architecture to achieve similar performance to more powerful schemes in the literature that exploit graph representations, for instance. However, the 2D joints/silhouettes estimates could be improved through the use of more accurate algorithms for the generation of ground truths. 
The VE did not require any form of pre-training, due to the fine-grained input from the SFE; this module was able to generate the parameters necessary to move from a 2D space to a 3D one, which is more easily applied in a diverse range of tasks and datasets.
It was important to experiment in order to find the right design for the VE, since this could have broken the entire system in scenarios of over- or underfitting where the produced viewpoint parameters do not allow for a correct estimation of the hand pose/shape. 
Finally, unlike other existing works, the HE was able to output accurate estimations without requiring an iterative process, as its re-projection procedure allowed for closer correlation between the 3D and 2D hand representations during training. However, a regularization term was still required, as without this the meshes completely collapsed onto themselves when the system tried to generate near-perfect spatial 3D joint estimations.
We further improved our architecture through the use of advanced strategies such as skeleton adaptation and hourglass output concatenation, to obtain both more refined 3D joint locations and finer grained input representations.
Our experimental results demonstrate that the multi-task SFE, VE, HE with weak re-projection, and the use of advanced strategies, which were designed by exploiting and extending schemes in the literature, achieved state-of-the-art performance on the task of 3D hand pose and shape estimation. 
Moreover, when applied to hand gesture recognition on both benchmark datasets, our framework outperformed other schemes that were devised for the estimation task and later employed to recognize hand gestures.

In future work, to address the current weaknesses of our system, we plan to upgrade the SFE by first increasing the accuracy of its 2D heatmaps and silhouette estimation by generating the corresponding ground truths via deep learning-based segmentation algorithms. We will also design other meaningful features to extend the multi-task strategy. 
In addition, we will explore solutions for the HE  that are not based on the MANO layer but on other approaches such as graph representations of the hand, in order to increase the abstraction capabilities of the model. In view of this, additional experiments will be performed in which we will retain the 3D shape when moving to the hand gesture recognition task, with the aim of improving the final results.
Although the proposed model currently addresses the pose/shape estimation of single hands by design, it could be extended to simultaneously handle inputs containing both hands. Thus, another possible avenue for future work would involve the exploration of alternative architectural extensions or design modifications to handle input images containing two hands.
Moreover, since the proposed architecture can classify roughly 27 images per second, we will design an adaptation to try to achieve real-time gesture recognition from video sequences.

\section*{Acknowledgment}
This work was supported in part by the MIUR under grant “Departments of Excellence 2018–2022” of the Department of Computer Science of Sapienza University. 

\bibliography{bibliography}

\end{document}